\documentclass[conference]{IEEEtran}
\IEEEoverridecommandlockouts
\usepackage{cite}
\usepackage{amsmath,amssymb,amsfonts}
\usepackage{algorithmic}
\usepackage{graphicx}
\usepackage{textcomp}
\usepackage{multirow}
\usepackage{xcolor}
\usepackage{caption}
\usepackage{makecell}
\usepackage{colortbl}
\usepackage{xcolor}
\usepackage{flushend}

\definecolor{rdcclight}{RGB}{235,245,255} % light blue (adjust if you want)
\def\BibTeX{{\rm B\kern-.05em{\sc i\kern-.025em b}\kern-.08em
    T\kern-.1667em\lower.7ex\hbox{E}\kern-.125emX}}

\definecolor{darkgreen}{rgb}{0,0.6,0}
\usepackage[bookmarks=true,colorlinks=true,pdfpagemode=UseNone,citecolor=darkgreen,linkcolor=blue,urlcolor=blue]{hyperref}

\begin{document}

\DeclareRobustCommand{\IChia}[1]{{\textcolor{blue}{#1}}} % I-Chia

\title{HybridMimic: Hybrid RL-Centroidal Control for Humanoid Motion Mimicking}
% \author{Anonymous Authors}
\author{
\IEEEauthorblockN{
Ludwig Chee-Ying Tay$^{{1}}$,
I-Chia Chang$^{{2}}$,
Yan Gu$^{2\dagger}$
}
\thanks{
$^{^{\dagger}}$Corresponding author.
$^{^{1}}$Ludwig Chee-Ying Tay is with Department of Computer Science, Purdue University, West Lafayette, IN 47907, USA.
E-mail: {\tt tay3@purdue.edu}.
$^{^{2}}$I-Chia Chang and Yan Gu are with the School of Mechanical Engineering, Purdue University, West Lafayette, IN 47907, USA. 
E-mail: {\tt \footnotesize \{chang970, yangu\}@purdue.edu.}
}
}

\maketitle

\begin{abstract}
% Motion mimicking, i.e., encouraging a control policy to imitate human movement, facilitates the learning of complex tasks via reinforcement learning (RL) for humanoid robots. 
% Although RL demonstrates great locomotion robustness and agility, standard RL frameworks do not explicitly reason about robot dynamics during deployment. 
% By integrating model-based methods, hybrid approaches have shown promise in improving locomotion control performance over standard RL.
% However, existing hybrid methods typically rely on predefined contact timing, which limits their application when contact timing is difficult to obtain, such as when mimicking diverse motions. 
% In this work, we introduce HybridMimic, a framework in which a learned policy predicts contact states and guides a centroidal-model-based controller to generate feedforward torques to imitate human motion. 
% Using physics-informed rewards, the policy learns to interact with the centroidal controller by predicting contact states, centroidal velocity commands, and reference torque.
% We demonstrate that HybridMimic provides a general framework for learning diverse, human-like motions by training it to mimic various motion clips. 
% Through sim-to-sim and sim-to-real experiments on the Booster T1 humanoid, HybridMimic reduces the average base position tracking error by 13\% compared to a state-of-the-art baseline.

Motion mimicking, i.e., encouraging the control policy to mimic human motion, facilitates the learning of complex tasks via reinforcement learning (RL) for humanoid robots. Although standard RL frameworks demonstrate impressive locomotion agility, they often bypass explicit reasoning about robot dynamics during deployment, which is a design choice that can lead to physically infeasible commands when the robot encounters out-of-distribution environments. By integrating model-based principles, hybrid approaches can improve performance; however, existing methods typically rely on predefined contact timing, limiting their versatility.
This paper introduces HybridMimic, a framework in which a learned policy dynamically modulates a centroidal-model-based controller by predicting continuous contact states and desired centroidal velocities. This architecture exploits the physical grounding of centroidal dynamics to generate feedforward torques that remain feasible even under domain shift. Using physics-informed rewards, the policy is trained to efficiently utilize the centroidal controller's optimization by outputting precise control targets and reference torques. Through hardware experiments on the Booster T1 humanoid, HybridMimic reduces the average base position tracking error by 13\% compared to a state-of-the-art RL baseline, demonstrating the robustness of dynamics-aware deployment.

\end{abstract}

\section{Introduction}

In the context of Reinforcement Learning (RL) for humanoids, motion mimicking, i.e., encouraging the control policy to mimic human motion via tracking rewards, serves as a technique to improve motion naturalness and sample efficiency while achieving locomotion and loco-manipulation tasks.
Motion mimicking has enabled humanoids to walk, run, and dance naturally \cite{Liao2025BeyondMimic, peng2018deepmimic} and perform loco-manipulation tasks, e.g., box picking \cite{Margolis2025SoftMimicLC}.
With these successes, it is promising to utilize motion mimicking to build embodied intelligence that can achieve diverse complex task for various applications. The primary objective of this work is to introduce HybridMimic, a hybrid control architecture that integrates RL with a model-based controller to enable humanoid robots to learn diverse and physically grounded motions. This framework addresses the substantial challenge of ensuring physical feasibility in standard RL during deployment while simultaneously eliminating the dependency of traditional model-based controllers on hand-crafted, predefined contact schedules.

% To mimic human motion, an RL policy is trained using physics simulators. Domain randomization \cite{Peng_2018, tobin2017domainrandomizationtransferringdeep} is used for robustness against a range of disturbances seen during the training process.
\subsection{Related Work}

\subsubsection{Standard RL}
RL policies are typically trained in physics-based simulators to ensure physical feasibility, and with domain randomization \cite{Peng_2018, tobin2017domainrandomizationtransferringdeep} employed to ensure robustness against various disturbances~\cite{gu2025evolution}.
However, standard RL frameworks, especially those that use a Proportional-Derivative (PD) controller to enable joint-level torque generation, do not explicitly reason about robot dynamics during deployment. 
As a result, when the policy's observations deviate from the training distribution, the policy network lacks a mechanism to generate physically feasible commands, leading to degraded sim-to-real performance.

\begin{figure}
    \centering
    \includegraphics[width=1.0\linewidth]{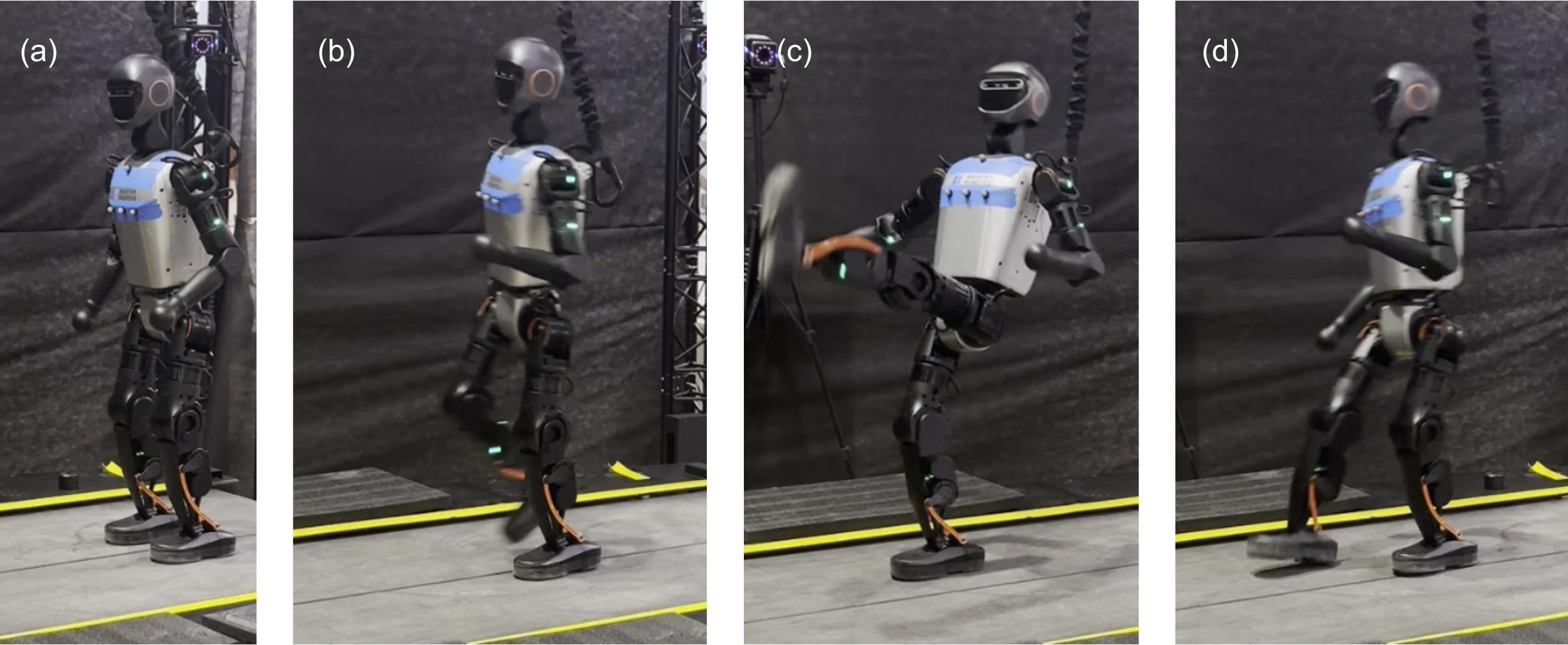}
    \caption{Snapshots of Booster T1 executing the kicking task using the proposed HybridMimic controller. (a) The robot starts standing. (b) The robot walks forward. (c) The robot kicks with its left foot. (d) The robot recovers and stabilizes from the kick and steps backwards. More motion can be found in the supplementary video. \url{https://youtu.be/1d5vkqNtCOY}}
    \label{fig: kick}
    % \vspace{-0.15 in}
\end{figure}

\subsubsection{Analytical-model-based control}
Despite their theoretical grounding~\cite{Westervelt2007} and impressive success in enabling dynamic locomotion in various environments~\cite{dosunmu2023stair,gao2025time,iqbal_provably_2020,dai2023data,kuindersma2016optimization}, model-based controllers face significant hurdles when tasked with enabling agile and robust locomotion in unstructured environments, due to the high-dimensional, hybrid, nonlinear nature of the associated robot dynamics. Classical inverted-pendulum-based methods~\cite{973365,gao2023time,xiong20223,stewart2025adaptive,iqbal2023mechatronics} often rely on simplifying assumptions, such as constant Center-of-Mass (CoM) height, which are frequently violated during natural human-like motions such as crouching or jumping. Even more sophisticated centroidal models~\cite{di2018dynamic} typically require predefined, rigid contact schedules to remain computationally tractable. These hand-crafted heuristics struggle to generalize to ``in-the-wild'' motions where contact timing is non-periodic or difficult to specify a priori. Furthermore, the performance of these controllers is strictly capped by the accuracy of the underlying reduced-order model; they lack the inherent ability to learn from experience or adapt to unmodeled environmental nuances. These limitations necessitate a learning-based approach that can discover optimal control strategies while still respecting fundamental physical constraints.

\subsubsection{Hybrid methods} Hybrid methods that integrate components or principles from model-based controllers into RL frameworks have been studied to improve performance.
To improve the training efficiency, \cite{Kang2023RLModelBased, yan2026efficiently} employ model-based outputs as reference signals to guide policy training.
Meanwhile, \cite{xie2022glide, cheng2025rambo, castillo2023template, Yang2023CAJun, egle2024enhancing} adopt model-based whole-body controllers as the torque generators in their RL frameworks.
These frameworks facilitate precise tracking \cite{cheng2025rambo}, higher walking speeds \cite{castillo2023template}, increased agility in quadruped jumping \cite{Yang2023CAJun}, and improved rejection of external disturbances \cite{egle2024enhancing}, demonstrating that the inclusion of a model-based torque controller can significantly improve locomotion performance.

While promising, existing hybrid methods rely on simplifying assumptions that limit their applicability for diverse motion mimicking. Inverted pendulum-based approaches \cite{castillo2023template, egle2024enhancing} assume a constant CoM height, an approximation that does not generally hold for human-like motion. Another common approach, centroidal methods \cite{cheng2025rambo, Yang2023CAJun}, typically require predefined footstep timings.

%While promising, a critical bottleneck remains: due to the difficulty of learning discrete decisions such as which end-effector is contacting using continuous learning methods, most existing hybrid architectures rely on hand crafted contact schedules to define step timings and simplify the underlying physics-based optimization. 
%This reliance on structured contact priors severely limits flexibility in motion mimicking, where contact transitions are highly dynamic and difficult to label a \textit{priori}, especially when the re-targeted motion is not accurate.

\subsection{Contributions}
In this paper, we propose HybridMimic, a hybrid control architecture that combines RL with a centroidal-model-based controller to generate feasible torque commands through centroid dynamics. Our HybridMimic formulation enables contact state estimation using policy observation, removing the requirement of hand-crafted contact schedules \cite{cheng2025rambo, Yang2023CAJun}. The proposed controller offers a general framework for learning diverse motions by training it to mimic human reference motions \cite{Liao2025BeyondMimic, peng2018deepmimic} (see Fig.\ref{fig: kick}). The specific contributions of this work are as follows:

%In this paper, we propose a novel framework, the contact schedule free HybridMimic composed of a hybrid of Reinforcement Learning and Model Based Optimal Control, to overcome these limitations. The specific contributions of this work are as follows:
\begin{enumerate}
%    \item \textbf{A Novel Hybrid Architecture}: 
%        We introduce HybridMimic, a framework where an RL policy serves as a high-level commander for a high-frequency centroidal controller to generate feedforward torques, while simultaneously providing joint-level PD targets for residual error correction. Our formulation allows the high-level policy to predict contact schedule, eliminating a key limiting assumption of previous centroidal policies \cite{Yang2023CAJun, cheng2025rambo}
        
%    We introduce HybridMimic, a framework where an RL policy serves as a high-level commander for a high-frequency centroidal controller to generate feedforward torques, while simultaneously providing joint-level PD targets for residual error correction. Our high-le

   \item [(a)] \textbf{Contact-schedule-free formulation}: 
Unlike existing hybrid centroidal methods \cite{Yang2023CAJun, cheng2025rambo}, HybridMimic estimates continuous contact states based on observations, eliminating the need for predefined contact schedules and providing smooth contact transitions. Additionally, HybridMimic optimizes the ground reaction forces by considering the reference torque output by the policy network. This allows diverse ground reaction force profiles to emerge naturally, making it well-suited to complex contact transitions inherent in motion mimicking.

   % \item \textbf{Contact-agnostic Formulation}: 
% Unlike existing hybrid centroidal methods \cite{Yang2023CAJun, cheng2025rambo}, HybridMimic evaluates ground reaction forces across all potential contact points and employs continuous weighting to zero non-contacting points, thereby enabling continuous, learned contact states. This formulation allows complex ground reaction force profiles to emerge naturally, making it well suited to complex contact transitions inherent in motion mimicking.

%    \item \textbf{Contact-Agnostic Formulation}: Unlike existing hybrid methods, HybridMimic does not require pre-defined contact schedules. It enables the emergence of arbitrary ground reaction force profiles, making it uniquely suited for the complex contact transitions inherent in motion mimicking.

    \item [(b)] \textbf{Physics-based rewards}: 
We introduce novel reward terms that are functions of the inputs and outputs of the centroidal controller, e.g., commanded accelerations, estimated contact states, and ground reaction forces. These rewards encourage consistent and physically grounded use of the centroidal controller by the policy, resulting in an interpretable and transparent model-based controller that can streamline physical deployment.

%     \item \textbf{Physics-based Rewards}: 
% We introduce novel reward terms using outputs of the centroidal controller, including centroidal accelerations, contact states, and ground reaction forces. These rewards encourage consistent and physically grounded use of the controller, resulting in an interpretable and transparent model-based controller which can streamline physical deployment.

%    \item \textbf{Improved Physical Interpretability}: With minimal reward design, the RL policy learns to utilize the centroidal controller in a physically meaningful manner. This transparency allows for more intuitive parameter tuning during training and physical deployment.
    \item [(c)] \textbf{Real-world deployment}:
Through extensive sim-to-sim and sim-to-real experiments, we demonstrate that HybridMimic successfully learns to perform diverse human motions while maintaining performance comparable to a state-of-the-art baseline \cite{Liao2025BeyondMimic}. We further highlight the benefits of integrating model-based control by showing a reduced sim-to-real gap in robot base position tracking during locomotion, achieving a 13\% reduction in base position error compared to the baseline.

\end{enumerate}
%Through extensive sim-to-sim and sim-to-real experiments, we show that HybridMimic significantly reduces tracking errors compared to state-of-the-art baselines, reducing base position tracking error by 17\%. To our knowledge, this represents the first successful deployment of such a hybrid approach for humanoid motion mimicking.

\section{Preliminaries}

\begin{figure*}
    \centering
    \includegraphics[width=0.99\textwidth]{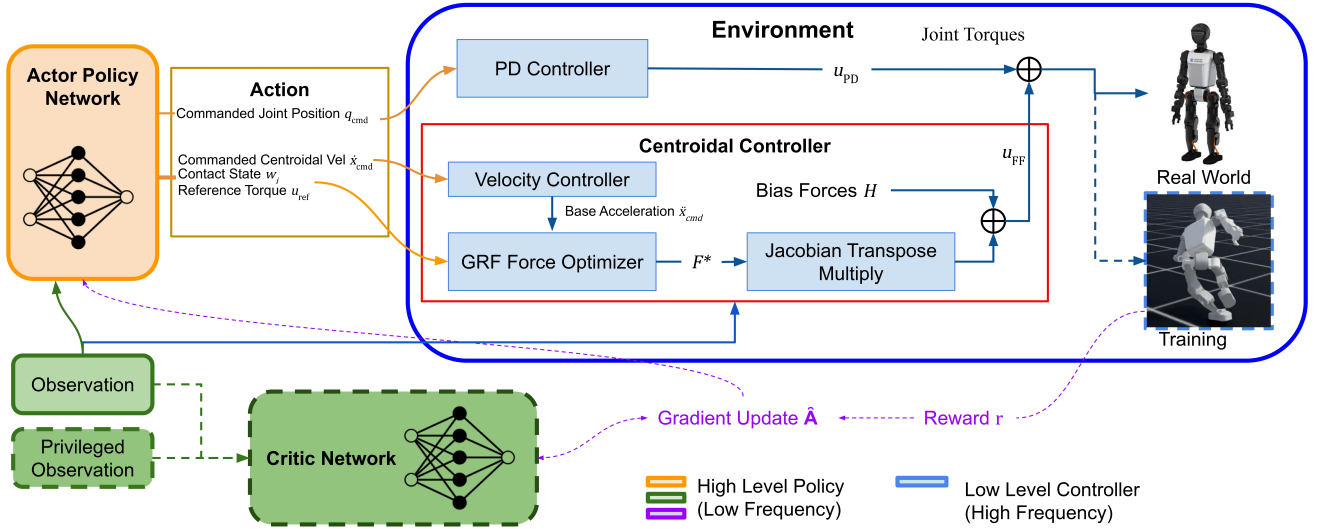}
    % \vspace{-0.2 in}
    \caption{Diagram illustrating HybridMimic controller formulation in both training and deployment. The dashed items represent training specific elements that are unused during deployment.}
    \label{fig:rdcc_diagr}
    % \vspace{-0.1 in}
\end{figure*}

We first review the robot dynamics models and the associated nomenclature for HybridMimic.
We focus on full-order dynamics to relate feedforward torque and ground reaction force, and centroidal dynamics to relate ground reaction force and centroidal acceleration.
These models bridge the high-level centroidal acceleration command to the low-level joint torque using physics-based models.

\subsection{Full-order Dynamics}

Using multi-rigid-body dynamics, the dynamics equation of a humanoid robot can be expressed as
\begin{equation}
    M(\tilde{q})\ddot{\tilde{q}}
    +
    H(\tilde{q},\dot{\tilde{q}})
    =
    B u
    +
    \sum_{i\in \mathcal{E}}
    J_i^\top F_i
    % \mathbf{M}(\mathbf{q})\ddot{\mathbf{q}}
    % +
    % \mathbf{H}(\mathbf{q},\dot{\mathbf{q}})
    % =
    % \mathbf{B} \mathbf{u}
    % +
    % \sum_{i\in E}
    % (\mathbf{J}_i)^\top \mathbf{F}_i
\end{equation}
where 
$\tilde{q} \in \mathbb{R}^{6+n}$ is the generalized coordinates of the floating-base humanoid robot with $n$ the total number of joints,
$M(\tilde{q})\in \mathbb{R}^{(6+n)\times(6+n)}$ is the inertial matrix,
$H(\tilde{q},\dot{\tilde{q}}) \in \mathbb{R}^{6+n}$ is the sum of Coriolis, centrifugal, and gravitational terms,
$u \in \mathbb{R}^{n}$ is the joint motor torque,
$B \in \mathbb{R}^{(n+6)\times n}$ is the input matrix, and
$F_i \in \mathbb{R}^{6}$ and $J_i  \in \mathbb{R}^{6 \times (n+6)}$ are the ground reaction wrench and the corresponding Jacobian for the contact indexed by $i \in \mathcal{E}$.
The set $\mathcal{E}$ represents the set of all possible contact surfaces on the robot.
In this work, a limited set $\mathcal{E} = \{\text{left foot}, \text{right foot} \}$ is considered. 
However, given feasible motion clips, the proposed framework could be readily extended to consider more possible contacting surfaces, e.g., hands, knees, and elbows.

\subsection{Centroidal Dynamics}

% $u_{FF}$, $F_{ang}$

% $u_{\text{FF}}$, $F_{\text{ang}}$

To enable high-frequency control and avoid complex computations during RL training, we use a single rigid body (SRB) model to capture the centroidal dynamics of the humanoid robot.
A SRB model assumes the mass of the robot is concentrated at the CoM, the angular inertia is constant, and the motion can be controlled by the ground reaction wrench $F_i$ with lever arms $r_i \in \mathbb{R}^{3}$ for all $i\in \mathcal{E}$ \cite{di2018dynamic}.
The linear acceleration $ \ddot{x}_{\text{lin}} \in \mathbb{R}^{3}$ and the angular acceleration $\ddot{x}_{\text{ang}} \in \mathbb{R}^{3}$ of the centroid can be expressed as
\begin{equation} \label{eq: c lin dyn}
m\ddot{{x}}_{lin} = mg + \sum_{i \in \mathcal{E}} F_{i,\text{lin}}
\end{equation}
\begin{equation} \label{eq: c ang dyn}
I_c \ddot{{x}}_{\text{ang}}
% = \sum_{i \in E} {F}_{i,\text{ang}} + \mathbf{r}^{(i)} \times {F}_{i, \text{lin}}
= \sum_{i \in \mathcal{E}} F_{i,\text{ang}} + [r_i]_{\times} F_{i,\text{lin}}
\end{equation}
where
$m \in \mathbb{R}$ is the total mass of the robot, 
$g \in \mathbb{R}^3$ is the gravitational acceleration vector,
$F_{i,\text{lin}} \in \mathbb{R}^3$ is the linear portion of $F_i$,
${I}_c \in \mathbb{R}^{3\times 3}$ is the rotational inertial matrix of the robot about the CoM,
$F_{i,\text{ang}} \in \mathbb{R}^3$ is the angular portion of $F_i$, and 
$[r_i]_{\times}$ is the skew-symmetric matrix of $r_i$.
Stacking the two equations above gives:
\begin{equation} \label{eq: full c}
\begin{aligned}
\begin{bmatrix}
\ddot{{x}}_{\text{lin}} \\
\ddot{{x}}_{\text{ang}}
\end{bmatrix}
=
\hat{{g}} 
+
\sum_{i \in \mathcal{E}} {A}_i F_i 
\end{aligned}
\end{equation}
where 
$\hat{{g}} = [{g}^\top,{0}_{1\times3}]^\top$,
${F}_i = [
    {F}_{i, \text{lin}}^\top,
    {F}_{i, \text{ang}}^\top ]^\top$
, and
$
   A_i = 
\begin{bmatrix}
\frac{1}{m}I_{3} 
& 
{0} 
\\
{I}_c^{-1}[{r}_i]_{\times}  
& 
{I}_c^{-1}
\end{bmatrix}
$.
The expression can be simplified as
\begin{equation}
    \ddot{{x}} = \hat{{g}} + {A} {F}
    \label{eq: SRBM}
\end{equation}
where 
$\ddot{{x}} = [\ddot{{x}}_{\text{lin}}^\top,
\ddot{{x}}_{\text{ang}}^\top ]^\top$,
${A}$ horizontally stacks all ${A}_i$ $ \forall i \in \mathcal{E}$, and
${F}$ vertically stacks all ${F}_i$ $ \forall i \in \mathcal{E}$.

\section{Methodology}
\label{se:Method}

% The goal of motion mimicking is to control a humanoid robot to mimic a motion demonstrated by a human.
% More specifically, a motion reference ${q}_{\text{ref}}(t)$ for $0<t<T_{m}$, where $T_m$ is the length of the motion clip, can be obtained by a motion retargeting method \cite{araujo2025retargeting}.
% Then, a controller needs to control the robot to minimize the difference between the motion reference ${q}_{\text{ref}}(t)$ and the actual motion $\mathbf{q}(t)$, as described by the tracking reward in Table~\ref{tab: Rewards}, by generating motor torque $\mathbf{u}(t)$.

To mimic a motion reference $\tilde{q}_{\text{ref}}(t)$ retargeted from a human demonstration \cite{araujo2025retargeting}, a controller needs to generate motor torque ${u}(t)$ to minimize the difference between the motion reference $\tilde{q}_{\text{ref}}(t)$ and the actual motion $\tilde{q}(t)$ performed by the robot.
For simplicity, we omit the explicit dependence on time in the rest of the paper.
Inspired by model-based control \cite{sadeghian2017passivity, reher2021inverse}, our HybridMimic controller (Fig.~\ref{fig:rdcc_diagr}) computes motor torque as the sum of a feedforward torque ${u}_{\text{FF}}$ and a PD torque ${u}_{\text{PD}}$:
\begin{equation}
    {u} = {u}_{\text{FF}} + {u}_{\text{PD}}
    \label{eq:hybrid controller}
\end{equation}
% Figure.~\ref{fig:rdcc_diagr} summarizes our overall HybridMimic formulation.
Similar to standard RL \cite{peng2018deepmimic ,Liao2025BeyondMimic}, the PD term is designed by
\begin{equation}
    {u}_{\text{PD}} = {K}_P ( {q}_{\text{cmd}}-{q})+{K}_D \dot{q}
    \label{eq:pd}
\end{equation}
where 
${K}_P$ and ${K}_D$ are the stiffness and damping matrices of motors, ${q} \in \mathbb{R}^{n}$ is the actual joint position, and ${q}_{\text{cmd}} \in \mathbb{R}^{n}$ is the commanded joint position generated by the policy network.
Using the PD term alone has been shown to be empirically effective for locomotion tasks and can implicitly handle both feedforward and feedback terms \cite{Lyu-RSS-20-P060}.
However, the PD formulation encodes all the control command into a single channel ${q}_{\text{cmd}}$ governed by the policy. 
Xiong et al. \cite{xiong2026extremcontrol} suggest that the standard PD formulation may cause latency without feedforward velocity command. 

Thus, in addition to the PD term, a centroidal controller is used to generate a feedforward torque ${u}_{\text{FF}}$ for HybridMimic.
% The key novelty of RDCC framework is the design of reward function to encourage the policy network to properly utilize the centroidal controller, improving the interpretability and enabling the policy to detect contact state without prior knowledge.
% In this section, we first introduce the computation of the feedforward torque. 
% Then, we present the policy training setup which encourages the policy to utilize the centroidal controller in an interpretable way.

\subsection{Feedforward torque generation}

To compensate for the effect of contact wrench and the bias force ${H}$, the feedforward torque is generated by the following:
\begin{equation}
    {u}_{\text{FF}} = -{B}^{\dagger}\sum_{i \in \mathcal{E}} J_i^\top {F}_i^{*} + {B}^{\dagger}{H}
    \label{eq:u_FF}
\end{equation}
where ${B}^{\dagger}$ is the pseudo-inverse of ${B}$ and ${F}_i^{*}$ is the feasible ground reaction wrench estimate for the contact $i\in \mathcal{E}$.

A feasible estimate of contacting wrenches ${F}_i^{*}$ $\forall i\in \mathcal{E}$ is obtained by solving the following constrained quadratic programming (QP) problem to respect the centroidal dynamics and the estimated contact states:
\begin{equation}
\begin{aligned}
\underset{{F}}{\text{min}} & \quad L({F}) \\
\text{subject to} & \quad {A} {F} = \ddot{{x}}_{\text{cmd}} - \hat{{g}}
\label{eq: qp prob}
\end{aligned}
\end{equation}
where the commanded centroidal acceleration $\ddot{{x}}_{\text{cmd}}$ is determined by
\begin{equation} \label{eq: k vel}
\ddot{{x}}_{\text{cmd}}
 = {K}_{\text{vel}}(\dot{{x}}_{\text{cmd}} - \dot{{x}})
\end{equation}
Here, ${K}_{\text{vel}} \in \mathbb{R}^{6\times6}$ is the gain for centroidal velocity tracking, $\dot{{x}}_{\text{cmd}} = [\dot{{x}}_{\text{cmd,lin}}^\top,\dot{{x}}_{\text{cmd,ang}}^\top]^\top \in \mathbb{R}^{6}$ is the commanded centroidal velocity output by the policy network, and $\dot{{x}} \in \mathbb{R}^{6}$ is the actual centroidal velocity. 
The cost function $L({F})$ is defined as
\begin{equation} \label{eq: QP Cost}
\begin{aligned}
L({F}) = &
\underbrace{\| {u}_{\text{ref}}
+\sum_{i \in \mathcal{E}}
J_i^\top F_i \|^2_{{K}_{\tau} \text{diag}({\tau}_{\text{limit}}^{-2})} }_{\text{Reference torque cost}}
\\
&+\underbrace{\sum_{i \in \mathcal{E}} 
\left\{ || {F}_{i, \text{lin}} ||^2_{e^{-w_i} {K}_{\text{lin}} }
+|| {F}_{i,\text{ang}} ||^2_{e^{-w_i}{K}_{\text{ang}}} \right\} }_{\text{End-effector contacting cost}}
\end{aligned}
\end{equation}
where 
${u}_{\text{ref}} \in \mathbb{R}^{n}$ is the reference torque generated by the policy,
${K}_{\tau}$ is the nominal weight of reference torque cost and is scaled by the motor torque limit 
${\tau}_{\text{limit}} \in \mathbb{R}^{n}$ where ${\tau}_{\text{limit}}^{-2}$ represents the element-wise inverse square of ${\tau}_{\text{limit}}$.
$w_i \in \mathbb{R}$ is the contact state for the $i^{\text{th}}$ contact surface in $\mathcal{E}$ and is generated by the policy. 
${K}_{\text{lin}}$ and ${K}_{\text{ang}}$ are the nominal weights of the end-effector contacting cost and are scaled by the contact state $w_i$. 
A larger $w_i$ implies that the $i^{\text{th}}$ surface is more likely to be in contact with the environment. We adopt a continuous description of contact state to ensure a continuous mapping from policy actions to joint torque, thereby avoiding sudden jumps in torque that may hinder RL training.
% $\mathbf{K}_{\tau}, {K}_{\text{lin}}, {K}_{\text{ang}}$ are the weight of each cost term with compatible dimension and are scaled as described below. 

% The weight of reference torque cost is scaled by the motor torque limit $\tau_{\text{limit}}$ to coordinate the strength of different motors.

\begin{table}
\caption{Actor and critic observations. Actor has access to the baseline observation while the critic has both the baseline and privileged observation. An observation history of four policy timesteps are stacked together. }
\centering
    \begin{tabular}{c|p{5cm}|c}
         \hline
         Term & \multicolumn{1}{c|}{Observation} & Dimension\\
         \hline
         \multicolumn{3}{c}{Baseline Observation}\\
         \hline
         ${q}_{\text{ref}}$ & Reference joint position & $n$\\
         $\dot{{q}}_{\text{ref}}$ & Reference joint velocity & $n$\\
         $\dot{{P}}_B$ & Base frame linear velocity & 3\\
         $\omega_B$ & Base frame angular velocity & 3\\
         ${q}$ & Current joint position & $n$\\
         $\dot{q}$ & Current joint velocity & $n$\\
         \hline
         \multicolumn{3}{c}{Privileged Observation}\\
         \hline
         ${p}_{\text{ref}}$ & Reference base position in world frame & 3\\
         ${p}$ & Current base position in world frame & 3\\
         ${R}_{WB, \text{ref}}$ & Reference base frame orientation relative to world frame (the 6 elements of the first two columns of a $3\times3$ Rotation matrix) & 6\\
         ${R}_{WB}$ & Current base frame orientation relative to world frame & 6\\
         \hline
    \end{tabular}
    % \vspace{-0.15 in}
    \label{tab:obs}
\end{table}

The cost function can be expressed as a quadratic form $L({F}) = \frac{1}{2}{F}^\top {Q} {F} + {c}^\top{F}$ where the detailed definitions of ${Q}$ and ${c}$ are omitted.
This allows the optimal solution to \eqref{eq: qp prob} to be expressed in a closed form as shown below, which facilitates the efficient computation during training and real-world deployment:
\begin{equation}
    {F}^*
    =
    - Q^{-1} c+ Q^{-1} A^\top\Big( A Q^{-1} A^\top\Big)^{-1}
    \Big(\ddot{{x}}-\hat{{g}}+ A Q^{-1} c\Big).
\end{equation}
Finally, the feedforward torque is clipped to respect the motor torque limit $\tau_{\text{limit}}$ to ensure safety and feasibility.

% The policy network controls the centroidal controller ground reaction wrench through reference torque ${u}_{\text{ref}}$ and contact state $w_i$. Both values are generated by the policy network. The contact state distinguishes between contacting and non-contacting end-effectors, enabling continuous reasoning over support conditions. The reference torque, applied in the absence of explicit friction cone and torque limit constraints, allows the policy to bias the resulting ground reaction wrench so that constraint-consistent behaviors emerge implicitly. We choose to represent the bias term as a reference joint torque, as previous works have demonstrated that RL can learn the complex joint torques necessary for locomotion \cite{Li2025SATA, Kim2023TorqueBasedDRL}

In our HybridMimic formulation, the policy network generates outputs to adjust the reference torque ${u}_{\text{ref}}$, the contact state $w_i$ for the $i^{\text{th}}$ contact, and the commanded centroidal velocity $\dot{x}_{\text{cmd}}$. These outputs enable the policy network to exploit the structure of the QP problem and the SRB dynamic model to generate feedforward torque without prior knowledge of the contact schedule. By scaling the end-effector contact cost weights using $w_i$, the formulation produces significant ground reaction wrenches only for contact surfaces with a high contact state value. Through the reference torque cost term, the policy network gains greater control over $F^{*}$, particularly when the equality constraint in \eqref{eq: qp prob} has infinitely many solutions (e.g., during the double-support phase), and implicitly prevents violations of the friction cone and actuation limit.

% In our HybridMimic formulation, the policy network generates output to control the reference torque ${u}_{\text{ref}}$, the contact state $w_i$ for the $i^{\text{th}}$ contact, and the commanded centroidal velocity $\dot{x}_{\text{cmd}}$.
% These outputs allow the policy network to exploit the structure provided by the QP problem and the SRB dynamic model to generate feedforward torque without prior knowledge of contact schedule. 
% By scaling the end-effector contacting cost weights using $w_i$, significant ground reaction wrenches are produced only for those end-effectors with a contact state value.
% Through the reference torque cost term, the policy network gains greater control of $\mathbf{F}$ when the equality constraint in \eqref{eq: qp prob} has infinitely many solutions (e.g., double support phase) and to implicitly prevent violation of friction cone and torque limit constraints.
%In addition, the tunable parameters, e.g., ${K}_{\text{vel}}$, allows parameter tuning to trade-off the sensor noise level and the tracking accuracy during deployment.
%However, the proper setup of the learning problem is essential to ensure the policy network to use the centroidal controller in a interatable way, which we introduce next.

\subsection{RL policy}

To train a policy network to utilize the proposed hybrid controller in \eqref{eq:hybrid controller}, an RL problem is formulated as a Partially Observable Markov Decision Process (POMDP),
represented by a tuple $\langle \mathcal{S}, \mathcal{O}, \mathcal{A}, \mathcal{P}, r, \gamma \rangle$. Here,
$\mathcal{S}$ is the state space,
$\mathcal{O}$ is the observation space (Tab.~\ref{tab:obs}),
$\mathcal{A}$ is the action space (including ${u}_{\text{ref}}$, $w_i$ $\forall i\in \mathcal{E}$, $\dot{x}_{\text{cmd}}$, and ${q}_{\text{cmd}}$),
$\mathcal{P}_{S' \mid S,A}$ is the state transition probability, 
$r : \mathcal{O} \times \mathcal{A} \rightarrow \mathbb{R}$ is the reward function (Tab.~\ref{tab: Rewards}), and
$\gamma$ is the discount factor. 
The policy $\pi_{\theta} : \mathcal{O} \rightarrow \mathcal{A}$, parameterized by $\theta$, is trained to maximize the expected discounted return over trajectories initialized from an initial state drawn from the initial distribution $s_0 \sim p_0$, where actions $a_t$ are selected given the observations $o_t$ at time step $t$:
\begin{equation}
\mathbb{E}_{s_0 \sim p_0(\cdot),\;
s_{t+1} \sim \mathcal{P}(\cdot \mid s_t, a_t),\;
a_t \sim \pi_{\theta}(\cdot \mid o_t)}
\left[
\sum_{t=0}^\top \gamma^\top \, r(s_t, a_t)
\right]
\end{equation}
We utilize Proximal Policy Optimization \cite{Schulman2017PPO} with an asymmetric actor-critic \cite{pinto2017asymmetric} for efficient learning to train one policy for each motion clip.
To improve robustness, domain randomization is used as summarized in Tab. \ref{tab: Domain Rand}.

\begin{table}
\caption{Values of reward terms. Reward terms are split into tracking, regularization, and HybridMimic categories. Tracking terms are taken directly from \cite{Liao2025BeyondMimic} with adjusted parameters. The HybridMimic terms are described in Secs. \ref{sec:grf}, \ref{sec:cs}, \ref{sec:tl}, and \ref{sec:acc}. For terms with sensitivity $\sigma$, the reward is defined as $r=e^{\frac{\epsilon}{\sigma^2}}$ where $\epsilon$ represents the error between the desired and measured values.
}
\begin{tabular}{c|c|c|c}
\hline
Type & Reward & Weight & Sensitivity \\
\hline
\multirow{6}{*}{Tracking} & Global anchor position & 0.7 & 0.3 \\
& Global anchor orientation & 0.7 & 0.4\\
& Body position & 1 & 0.2\\
& Body orientation & 1 & 0.3\\
& Body linear velocity & 1 & 0.5\\
& Body angular velocity & 1.25 & 2.09\\
\hline
\multirow{2}{*}{Regularization} & L2 action rate & -0.08 & - \\
& Undesired contacts & -0.5 & -\\
\hline
\multirow{4}{*}{HybridMimic} & GRF reward & 0.15 & 100 \\
& Contact state reward & 0.5 & -\\
& Torque limit reward & 0.1 & -\\
& Centroidal acceleration (lin) & 0.1 & 6\\
& Centroidal acceleration (ang) & 0.1 & 12\\
\hline
\end{tabular}
\label{tab: Rewards}
    % \vspace{-0.2 in}
\end{table}

\begin{table}
\caption{Settings of domain randomization applied to improve policy robustness. Terms with + have a uniformly distributed random noise added to the true value. Terms with * have a uniformly distributed random factor multiplied to the true value. We add random noise to HybridMimic inputs to prevent the policy from learning behaviors that exploit using the centroidal controller for noise-free observation.}
\begin{tabular}{c|c|c}
\hline
Type & Randomization Term & Range \\
\hline
\multirow{4}{*}{Observations} & Base linear velocity noise (m/s) & +(-0.5, 0.5)\\
& Base angular velocity noise (rad/s) & +(-0.2, 0.2)\\
& Joint position noise (rad) & +(-0.02, 0.02)\\
& Joint velocity noise (rad/s) & +(-0.5, 0.5)\\
\hline
\multirow{4}{*}{\makecell{HybridMimic\\input}} & Mass randomization $M$ (kg) & *(0.98, 1.02)\\
& Inertia matrix ${I}_c$ & *(0.98, 1.02) \\
& Jacobian ${J}$ & *(0.98, 1.02) \\
& End-effector position ${r}$ (m) & +(-0.02, 0.02)\\
\hline
\multirow{3}{*}{Environment} & Initial joint position (rad)& +(-0.1, 0.1)\\
& Body CoM position offset (m)& +(-0.025, 0.025)\\
& Base velocity push (m/s) & +(1.0, 3.0)\\
\hline
\end{tabular}

\label{tab: Domain Rand}
% \vspace{-0.15 in}
\end{table}

A key novelty of this work is the introduction of reward functions that ensure feasible utilization of the centroidal controller. We encourage feasible utilization of the controller by minimizing the error between simulated values, such as ground reaction wrench ${F}_{\text{sim}}$, and the predicted value, ${F}^*$, through reward design. Our rewards facilitate the derivation of accurate, interpretable outputs, including ground reaction wrenches, contact states, and centroidal acceleration.

The following sections describe the specific rewards implemented for HybridMimic.

\subsubsection{Ground reaction force (GRF) reward}
\label{sec:grf}
The policy network should, when utilizing the centroidal controller correctly, generate an estimate of ground reaction wrench ${F}_{i}^{*}$ close to the value ${F}_{i,\text{sim}}$ from the simulator for the $i^{\text{th}}$ contact. 
Thus, a reward is given for minimizing the error between the two values. 
\begin{equation}
r_{\text{GRF}} = 
\exp \bigg(-\frac{\sum_{i \in \mathcal{E}}\|{F}_{i,\text{sim}} - {F_i^*}\|^2} {\sigma_{grf}^2}\bigg)
\end{equation}
The sensitivity constant $\sigma_{grf} \in \mathbb{R}$ can be chosen based on the magnitude of the expected contacting wrench of the humanoid robot.

\subsubsection{Contact state reward}
\label{sec:cs}
The contact state $w_i$ generated by the policy network should match the contact state $w_{i,\text{sim}}$ obtained through the simulator. 
A penalty is given for the predicted contact state of the policy network deviating from the true contact state
\begin{equation}
    r_{\text{contact}} = -\sum_{i \in \mathcal{E}}(\text{sigmoid}(w_{i,\text{sim}}) - \text{sigmoid}(w_i))^2
\end{equation}

\subsubsection{Torque limit reward}
\label{sec:tl}
For computational efficiency, the centroidal controller does not consider torque limits.
The policy network is intended to implicitly learn to utilize ${u}_{\text{ref}}$ to avoid exceeding torque limits. 
A torque limit penalty is applied when ${u}_{\text{FF}}$ exceeds a soft torque limit.
\begin{equation}
    r_{\tau_\text{limit}} = -
    \sum_{j = 1}^{n}  \max(0,{u}_{\text{FF},j} - \alpha {\tau}_{\text{limit},j})
\end{equation}
where $\alpha \approx 0.9$ and the subscript ${[\cdot]}_{j}$ indicates the $j^{\text{th}}$ joint.

\subsubsection{Centroidal acceleration reward}
\label{sec:acc}
The robot's acceleration in simulation should match the commanded acceleration computed by equation~\eqref{eq: k vel}. 
To encourage this behavior, a centroidal acceleration reward is implemented for minimizing the error between the simulated acceleration $\ddot{{x}}_{\text{sim}}$ and the commanded centroidal acceleration $\ddot{{x}}_{\text{cmd}}$.
\begin{equation}
    r_{acc} = \exp\bigg(-\frac{\|\ddot{{x}}_{\text{cmd}} - \ddot{{x}}_{\text{sim}}\|^2}{\sigma_{acc}^2}\bigg)
\end{equation}
where $\sigma_{acc} \in \mathbb{R}$ is the sensitivity constant detailed in Table~\ref{tab: Rewards}.

% \clearpage

\section{Results and Analysis}

To validate the effectiveness of the proposed HybridMimic framework to mimic human motion, we conduct an extensive study of both sim-to-sim and sim-to-real transfer. 
In the study, several variations of the controller are compared to show that the proposed design is crucial to achieving the improved motion mimicking performance.
The controllers include:
\begin{itemize}
    \item [(i)] \textbf{BeyondMimic}~\cite{Liao2025BeyondMimic}: This produces an RL policy using only a PD controller \eqref{eq:pd} as the torque generator, and
    serves as our baseline to highlight the performance gains introduced by the centroidal controller.
    
    \item [(ii)] \textbf{HybridMimic}: The proposed controller from Sec. \ref{se:Method}.

    \item [(iii)] \textbf{HybridMimic with fixed contact schedule (FCS) (termed as ``HybridMimic+FCS'')}: This is HybridMimic but with the contact state $w_i$ derived from the reference motion. 
    We set $w_i = 1$ if the distance between the $i^{\text{th}}$ contact surface and the environment is below a threshold and $w_i = 0$ otherwise.
    This ablation is used to demonstrate that the policy-estimated contact states $w_i$ are necessary for robust motion mimicking.
    
    \item [(iv)] \textbf{HybridMimic+FCS without Reference Torque Cost (RTC) (termed as ``HybridMimic+FCS-RTC'')}: This is the HybridMimic+FCS formulation with the reference torque cost removed from the cost function $L({F})$. 
    As our most complete ablation, this setup closely mirrors previous hybrid centroidal controllers \cite{cheng2025rambo,Yang2023CAJun}.
    We use it to show that policy-generated contact states $w_i$ and reference torques ${u}_{\text{ref}}$ are essential for robust performance.
    
\end{itemize}
% BeyondMimic \cite{Liao2025BeyondMimic} is used as a baseline to show the improvement caused by the centroidal controller.
% HybridMimic+FCS and HybridMimic+FCS-RTC are used to show that the contact state $w_i$ and reference torque ${u}_{\text{ref}}$ predicted by policy network are necessary to learn a robust controller for motion mimicking.

To evaluate the performance of a controller, we compare the difference between the trajectory rollout in the training environment ${q}_{\text{train}}$ (which is a good approximation of ${q}_{\text{ref}}$ after sufficient training) and the rollout during evaluation ${q}_{\text{eval}}$ either in the MuJoCo simulation or in the real experiment.
We use ${q}_{\text{train}}$ as a baseline instead of the motion reference ${q}_{\text{ref}}$ since the latter may not be dynamically feasible to be tracked by the robot due to retargeting issues \cite{araujo2025retargeting}. 
The difference between ${q}_{\text{eval}}$ and ${q}_{\text{train}}$ also characterizes the capability of the controller under domain shift caused by sim-to-sim or sim-to-real transfer. 
In this study, the following performance metrics are used: 
\begin{itemize}
    \item [(a)] \textbf{Base final position error (m)}: The L2 norm of the base position error at the end of the motion clip.
    \item [(b)] \textbf{Base mean position error (m)}: The averaged L2 norm of the base position error across the entire motion clip.
    \item [(c)] \textbf{Base mean orientation error (rad)}: The averaged L2 norm of orientation error across the entire motion clip.
    \item [(d)] \textbf{Base velocity root-mean-square (RMS) error (m/s)}: The RMS value of the L2 norm of linear velocity error for the entire motion clip.
    \item [(e)] \textbf{Base angular velocity RMS Error (rad/s)}: The RMS value of the L2 norm of angular velocity error for the entire motion clip.
\end{itemize}

\subsection{Training}
We implemented HybridMimic in IsaacLab \cite{IsaacLab2024} and trained our policies on a Booster T1 humanoid robot for various motion clips.  
The parameters used in the training environment are described in Tab. \ref{tab:Train}. 
The policy network is operated at 50 Hz and the centroidal controller runs at 500 Hz. 
% Note that the high frequency control loop frequency is changed for sim-to-real deployment due to hardware limitations. 
% We found that the HybridMimic operates with lower joint stiffness than the vanilla PD controller. 
The stiffness matrix ${K}_{P}$ for HybridMimic is 0.75 times that suggested in \cite{Liao2025BeyondMimic}.
When training the policy on a walking task using a partial A100 GPU instance provided by \cite{boerner2023access}, we measured the wall clock time to be 15.4 hours for BeyondMimic and 22.3 hours for HybridMimic, representing a computational overhead of $31\%$ in training time.

\begin{table}[t]
\caption{Training and HybridMimic parameters.}
\centering
\begin{tabular}{cc|cc}
\hline
\multicolumn{4}{c}{Training Parameters}\\
\hline
Term & Value & Term & Value\\
\hline
\# of environments & 8192 & \# of iterations & 15000\\
Episode length & 10s & Learning rate & $10^{-3}$\\
Desired KL & 0.01 & Value loss coeff. & 1.0\\
Entropy coeff. & 0.005 & Clip ratio & 0.2\\
Actor size & [1024,512,256,256] & $\gamma$ (Discount factor) & 0.99 \\
Critic size & [1024,512,256,256] & $\lambda$ (GAE decay) & 0.95\\
\hline
\multicolumn{4}{c}{HybridMimic Parameters}\\
\hline
Term & Value & Term & Value\\
\hline
${K}_{\text{vel,lin}}$ & $5I_3$ & ${K}_{\text{vel,ang}}$ & $10I_3$\\
${K}_{\tau}$ & $100I_{n}$ 
& ${K}_{\text{lin}}$ & $10^{-3}I_3$\\
${K}_{\text{ang}}$ & $0.02I_3$\\
\hline
\end{tabular}
\label{tab:Train}
% \vspace{-0.2 in}
\end{table}

\subsection{Sim-to-Sim Results}

\begin{table*}[t]
\centering
\caption{Sim-to-sim and sim-to-real tracking errors across motion clips and controller variants. 
The bolded figures represent the \textbf{lowest} tracking error in the category. 
The underlined figures represent the \underline{second best} in the category.
The text with background color is the \colorbox{rdcclight}{proposed method}.
}
\begin{tabular}{p{1.3cm}|p{2.9cm}|p{2.2cm}|p{2.2cm}|p{2.2cm}|p{2.2cm}|p{2.2cm}}
\hline
\multicolumn{7}{c}{Sim-to-Sim Results}\\
\hline
\multicolumn{1}{c|}{Task} & \multicolumn{1}{c|}{Model} & Base Final Position Error (m) & Base Mean Position Error (m) & Base Mean Orientation Error (rad) & Base Velocity Error (m/s) & Base Angular Velocity Error (rad/s)\\
\hline

\multirow{4}{*}{\makecell{Walking \\ to Kneeling}}
& BeyondMimic
& 0.4739 $\pm$ 0.031
& 0.3344 $\pm$ 0.021
& 0.1577 $\pm$ 0.012
& \underline{0.1331 $\pm$ 0.004}
& \underline{0.2881 $\pm$ 0.007}\\

& HybridMimic+FCS
& 0.6824 $\pm$ 0.021
& 0.4768 $\pm$ 0.014
& 0.2356 $\pm$ 0.007
& 0.1291 $\pm$ 0.002
& 0.3021 $\pm$ 0.004\\

& HybridMimic+FCS-RTC
& \underline{0.3641 $\pm$ 0.015}
& \underline{0.2633 $\pm$ 0.013}
& \textbf{0.1513 $\pm$ 0.006}
& 0.1202 $\pm$ 0.004
& 0.2679 $\pm$ 0.007\\

& \cellcolor{rdcclight} HybridMimic
& \cellcolor{rdcclight} \textbf{0.3455 $\pm$ 0.011}
& \cellcolor{rdcclight} \textbf{0.2366 $\pm$ 0.008}
& \cellcolor{rdcclight} \underline{0.1576 $\pm$ 0.003}
& \cellcolor{rdcclight} \textbf{0.1105 $\pm$ 0.001}
& \cellcolor{rdcclight} \textbf{0.2285 $\pm$ 0.002}\\

\hline
\multirow{4}{*}{\makecell{Running \\in Circle}}
& BeyondMimic
& \underline{0.3435 $\pm$ 0.008}
& \underline{0.2653 $\pm$ 0.007}
& \underline{0.1597 $\pm$ 0.005}
& \underline{0.1985 $\pm$ 0.002}
& \underline{0.3961 $\pm$ 0.002}\\

& HybridMimic+FCS
& 0.5626 $\pm$ 0.042
& 0.2930 $\pm$ 0.031
& 0.1757 $\pm$ 0.013
& 0.2066 $\pm$ 0.003
& 0.5064 $\pm$ 0.004\\

& HybridMimic+FCS-RTC
& 1.8963 $\pm$ 0.009
& 0.7606 $\pm$ 0.009
& 0.4071 $\pm$ 0.005
& 0.4894 $\pm$ 0.000
& 0.8870 $\pm$ 0.001\\

& \cellcolor{rdcclight} HybridMimic
& \cellcolor{rdcclight} \textbf{0.3289 $\pm$ 0.012}
& \cellcolor{rdcclight} \textbf{0.2622 $\pm$ 0.010}
& \cellcolor{rdcclight} \textbf{0.1413 $\pm$ 0.009}
& \cellcolor{rdcclight} \textbf{0.1497 $\pm$ 0.001}
& \cellcolor{rdcclight} \textbf{0.3711 $\pm$ 0.004}\\

\hline
\multirow{4}{*}{Jumping}
& BeyondMimic
& 0.4670 $\pm$ 0.018
& 0.2794 $\pm$ 0.010
& 0.2827 $\pm$ 0.006
& 0.5346 $\pm$ 0.001
& 1.4305 $\pm$ 0.002\\

& HybridMimic+FCS
& \underline{0.3771 $\pm$ 0.012}
& \underline{0.2225 $\pm$ 0.007}
& \underline{0.1518 $\pm$ 0.004}
& \underline{0.2168 $\pm$ 0.001}
& \underline{0.5657 $\pm$ 0.002}\\

& HybridMimic+FCS-RTC
& \textbf{0.3571 $\pm$ 0.009}
& \textbf{0.1929 $\pm$ 0.006}
& \textbf{0.1253 $\pm$ 0.004}
& \textbf{0.1957 $\pm$ 0.000}
& \textbf{0.4566 $\pm$ 0.002}\\

& \cellcolor{rdcclight} HybridMimic
& \cellcolor{rdcclight} 0.3926 $\pm$ 0.017
& \cellcolor{rdcclight} 0.2217 $\pm$ 0.009
& \cellcolor{rdcclight} 0.1521 $\pm$ 0.005
& \cellcolor{rdcclight} 0.2306 $\pm$ 0.002
& \cellcolor{rdcclight} 0.5548 $\pm$ 0.003\\

\hline
\multicolumn{7}{c}{Sim-to-Real Results} \\
\hline

\multirow{2}{*}{Sidestepping}
& BeyondMimic
& \underline{0.0942 $\pm$ 0.042}
& \underline{0.0471 $\pm$ 0.003}
& \textbf{0.0614 $\pm$ 0.004}
& \underline{0.2767 $\pm$ 0.092}
& \textbf{0.7708 $\pm$ 0.006}\\

& \cellcolor{rdcclight} HybridMimic
& \cellcolor{rdcclight} \textbf{0.0694 $\pm$ 0.008}
& \cellcolor{rdcclight} \textbf{0.0401 $\pm$ 0.002}
& \cellcolor{rdcclight} \underline{0.0962 $\pm$ 0.001}
& \cellcolor{rdcclight} \textbf{0.2676 $\pm$ 0.078}
& \cellcolor{rdcclight} \underline{0.7810 $\pm$ 0.001}\\

\hline
\multirow{2}{*}{\makecell{Backwards\\Stepping}}
& BeyondMimic
& \underline{0.1849 $\pm$ 0.002}
& \underline{0.0500 $\pm$ 0.001}
& \underline{0.0912 $\pm$ 0.002}
& \underline{0.1994 $\pm$ 0.001}
& \underline{0.9044 $\pm$ 0.003}\\

& \cellcolor{rdcclight} HybridMimic
& \cellcolor{rdcclight} \textbf{0.1079 $\pm$ 0.011}
& \cellcolor{rdcclight} \textbf{0.0405 $\pm$ 0.002}
& \cellcolor{rdcclight} \textbf{0.0834 $\pm$ 0.004}
& \cellcolor{rdcclight} \textbf{0.1947 $\pm$ 0.005}
& \cellcolor{rdcclight} \textbf{0.8978 $\pm$ 0.003}\\

\hline
\multirow{2}{*}{\makecell{Forwards\\Walking}}
& BeyondMimic
& \underline{0.2275 $\pm$ 0.067}
& \underline{0.0592 $\pm$ 0.006}
& \textbf{0.0808 $\pm$ 0.011}
& \underline{0.3293 $\pm$ 0.036}
& \underline{0.8762 $\pm$ 0.078}\\

& \cellcolor{rdcclight} HybridMimic
& \cellcolor{rdcclight} \textbf{0.1377 $\pm$ 0.025}
& \cellcolor{rdcclight} \textbf{0.0496 $\pm$ 0.001}
& \cellcolor{rdcclight} \underline{0.0912 $\pm$ 0.006}
& \cellcolor{rdcclight} \textbf{0.1856 $\pm$ 0.004}
& \cellcolor{rdcclight} \textbf{0.7581 $\pm$ 0.001}\\
\hline
\multirow{2}{*}{Kicking} & BeyondMimic & \underline{0.1564 $\pm$ 0.051} & \underline{0.0580 $\pm$ 0.001} & \textbf{0.1338 $\pm$ 0.007} & \underline{0.9608 $\pm$ 0.016} & \underline{0.6388 $\pm$ 0.014}\\
 & \cellcolor{rdcclight} HybridMimic 
 & \cellcolor{rdcclight} \textbf{0.1131 $\pm$ 0.009}
 & \cellcolor{rdcclight} \textbf{0.0567 $\pm$ 0.002}
 & \cellcolor{rdcclight} \underline{0.1826 $\pm$ 0.005} 
 & \cellcolor{rdcclight} \textbf{0.9402 $\pm$ 0.008} 
 & \cellcolor{rdcclight} \textbf{0.6360 $\pm$ 0.006}\\

\hline
\end{tabular}

\label{tab:sim2sim_tracking_stats}
    % \vspace{-0.15 in}
\end{table*}

We train all four policies on three 10-$s$ motion clips: 
\begin{itemize}
    \item [(1)] \textbf{Walking to kneeling}: Robot walks forward then high kneels on the right leg \cite{accad_mocap_data}.
    \item [(2)] \textbf{Running in Circle}: Robot runs in a circle \cite{cmu_mocap_database}.
    \item [(3)] \textbf{Jumping}: Robot jumps forward using both legs \cite{harvey2020robust}.
\end{itemize}
We evaluated the trained policies in the IsaacLab training environment without domain randomization to obtain datasets of 256 trained trajectories ${q}_{\text{train}}$, with a large count to mitigate the nondeterministic effects of the IsaacLab environments. 
We then evaluate the policies in MuJoCo to obtain ${q}_{\text{eval}}$. The sim-to-sim tracking results are shown in Tab. \ref{tab:sim2sim_tracking_stats}, and the video can be found in the supplementary video.

For all three motion clips, HybridMimic and BeyondMimic are able to mimic the reference motion, while HybridMimic demonstrates lower tracking error in linear position, linear velocity, and angular velocity, showing the proposed framework improves the motion mimicking performance through the feedforward torque.
By comparing the HybridMimic with its variants, it can be found that the HybridMimic+FCS and HybridMimic+FCS-RTC have higher tracking error than full HybridMimic for the first two motion clips, where the contact timing is more complex than the jumping motion.
The strong performance of HybridMimic+FCS-RTC in the jumping task likely stems from the motion’s simple contact schedule, which allows for easier learning due to the lack of redundant action space in full HybridMimic. However, the fragility of fixed contact schedules becomes evident in running, where complex contact patterns and disturbances require adaptable scheduling.
% The sim-to-sim results suggest that HybridMimic controls the base position better than the baseline BeyondMimic control over diverse tasks.

\subsection{Hardware Experiments}
We train the HybridMimic and BeyondMimic policies on a series of motion clips involving dynamic locomotion tasks:
\begin{itemize}
    \item [(1)] \textbf{Forward walking}: Starting from standing, the robot walks forward \cite{cg-2007-2}. The task has duration $7s$.
    \item [(2)] \textbf{Side-stepping}: Sidestep to the right \cite{cmu_mocap_database}. The task has duration $6.5s$.
    \item [(3)] \textbf{Backwards stepping}: Stepping backwards \cite{cmu_mocap_database}. The task has duration $6s$.
    \item [(4)] \textbf{Kicking}: Robot steps forward then kicks with the left leg before stepping backwards to original location \cite{cg-2007-2}. The task has duration $8s$.
\end{itemize}
The trained policies are deployed on the real robot. VICON motion-capture cameras are used both to capture tracking data and to provide linear velocity information for policy observation at 100 Hz. On the real robot, the centroidal controller was run at 350 Hz due to hardware limitations.
Each policy was executed to obtain three successful trials per motion clip. HybridMimic performs complex dynamic tasks in the real world (Fig. \ref{fig: kick}, supplementary video), with tracking statistics shown in Tab. \ref{tab:sim2sim_tracking_stats}.

\begin{figure}
    \centering
    \includegraphics[width=1.0\linewidth]{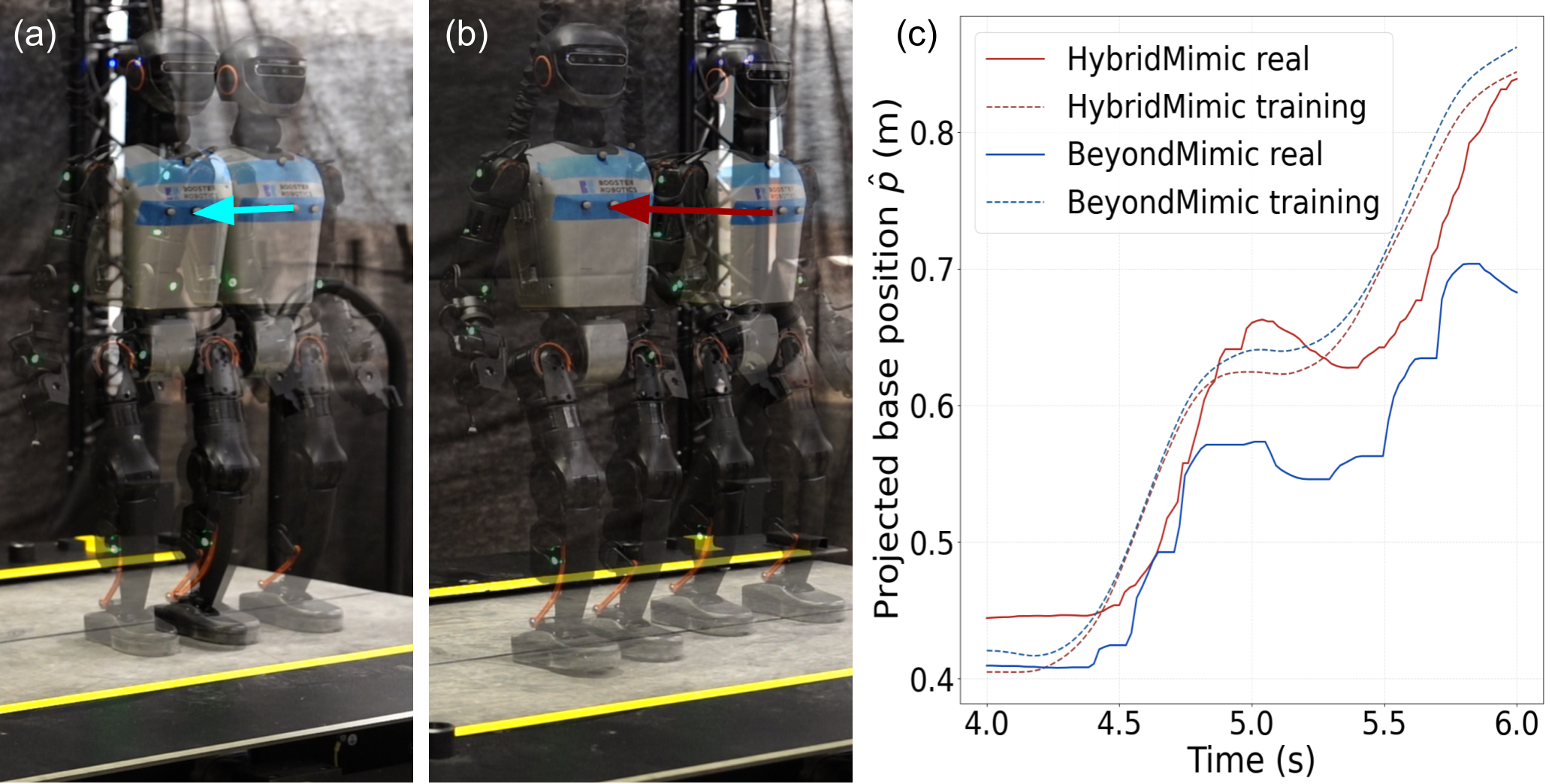}
    \caption{
    The left two images illustrate the robot displacement after two sidesteps for (a) BeyondMimic and (b) HybridMimic controllers.
    (c) The projected base position $\hat{p}$ is the base position projected onto the line from the starting to ending position in the reference motion clip. While the real HybridMimic trajectory oscillates around the trained motion, the real BeyondMimic trajectory consistently undershoots it and induces trunk wobbling as shown in the experiment video.}
    \label{fig:1d}
        % \vspace{-0.2 in}
\end{figure}

As shown in Tab. \ref{tab:sim2sim_tracking_stats}, HybridMimic achieves an about $13\%$ reduction in base position tracking error across all four real-world motions compared to the BeyondMimic baseline.
This demonstrates that the proposed HybridMimic improves the sim-to-real tracking performance compared to a state-of-the-art baseline.
For the sidestepping motion, Fig. \ref{fig:1d} shows the robot's base position trajectory in the direction of travel.
As shown in Fig. \ref{fig:1d}, the proposed HybridMimic performs a smoother motion and stays closer to the trajectory shown in training environments, while the motion controlled by BeyondMimic is more jittery and has a larger base position tracking error and can be illustrated in the experiment video.

During sim-to-real deployment, we noted that tuning HybridMimic parameters was comparatively straightforward due to the transparent and explainable nature of the proposed design.
One example involves adjusting the gain ${K}_{\text{vel}}$ for centroidal velocity tracking to balance the sensor noise level and the velocity tracking performance.

\subsection{Results Analysis}
Alongside demonstrating the reduced tracking error, another feature of HybridMimic is the transparent and interpretable behavior of the centroidal controller. 
In this section, we discuss the ground reaction force solved from the QP problem \eqref{eq: qp prob} and the feedforward torque generated by HybridMimic and its variant during the walking then kneeling motion task in a sim-to-sim study.

\begin{figure}
    \centering
    \includegraphics[width=1.0\linewidth]{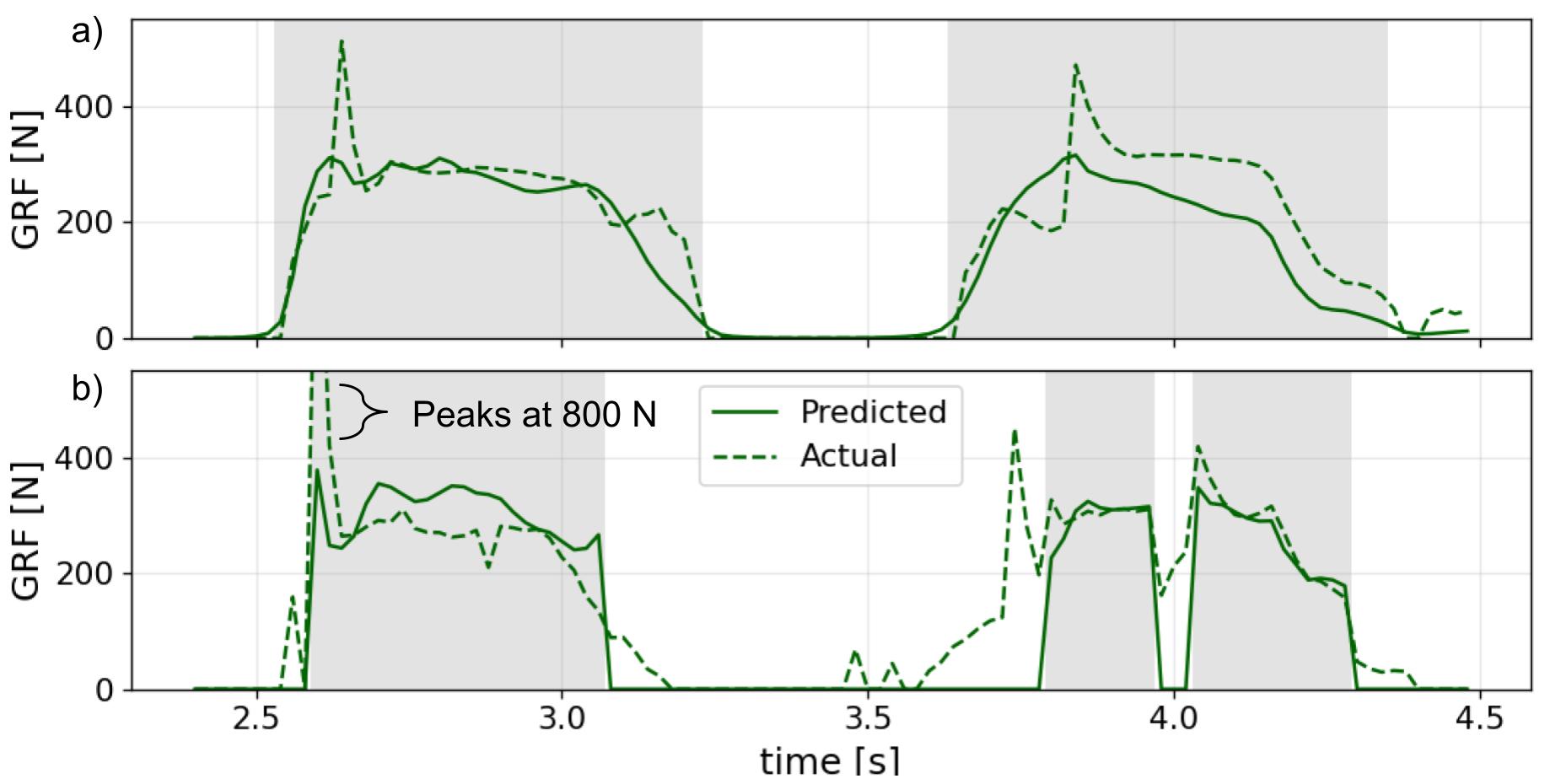}
    \caption{The vertical ground reaction force of left foot during walking in simulation for (a) HybridMimic and (b) HybridMimic+FCS.
    The highlighted regions indicate periods of time during which the contact state indicates the left foot is contacting the ground.
    For HybridMimic, the highlighting occurs when $w_{\text{left foot}} >0$ and for HybridMimic+FCS , the highlighting occurs when the reference motion has a foot height below 0.007m.}
    \label{fig:grf}
    % \vspace{-0.2 in}
\end{figure}

The estimated ground reaction forces of HybridMimic and HybridMimic+FCS are shown in Fig. \ref{fig:grf}. HybridMimic learns to accurately estimate ground reaction forces, with a small error between the simulated and estimated ground reaction force. Visual illustration can be found in the supplementary video.
Additionally, we compare it to HybridMimic+FCS and show that learning contact states is crucial because errors in the contact schedule derived from the motion reference lead to brittle and inaccurate ground reaction force computation.
This downgrades the tracking performance through inaccurate feedforward torque generation in \eqref{eq:u_FF} and requires the policy to generate additional PD torque to compensate for it.

\begin{figure}
    \centering
    \includegraphics[width=1.0\linewidth]{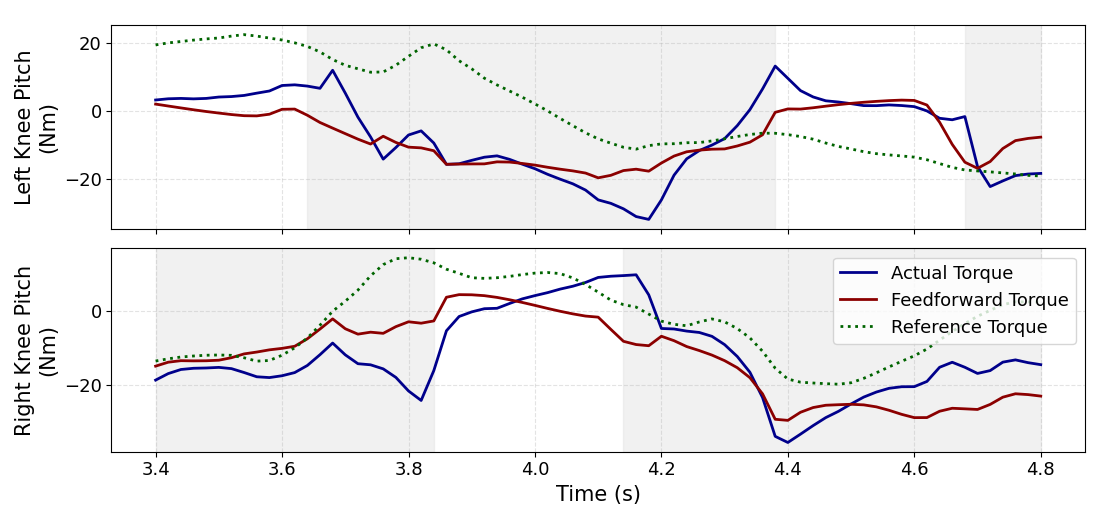}
    \caption{Feedforward, actual, and reference torque during walking for the left and right knee pitch joints.
    The highlighted region corresponds to contact on the corresponding leg determined by when $w_i > 0$.}
    \label{fig:u_ff}
        % \vspace{-0.2 in}
\end{figure}

Figure \ref{fig:u_ff} shows the feedforward torque ${u}_{\text{FF}}$, reference torque ${u}{\text{ref}}$, and simulated torque for the left and right knee joints under HybridMimic. When a foot contacts the ground, the actual and feedforward torques are similar, indicating that most motor torque is generated by the centroidal-controller-driven feedforward term. During swing phase, the feedforward torque remains small but nonzero due to the ${H}$ term in \eqref{eq:u_FF}. The reference torque roughly follows the feedforward torque but is not expected to match it, as it is an additional control channel for the policy.

\section{Conclusion and Future Work}
This paper has introduced HybridMimic, a hybrid framework integrating centroidal-model-based control with reinforcement learning (RL) to track diverse, dynamic motion references.
% Through physics-informed reward design, the policy network learns to generate high-level commands and estimate contact states to utilize the centroidal controller to generate feedforward torque, effectively counteracting the ground contact wrenches.
Through physics-informed reward design, the policy network learns to modulate commanded centroidal velocities and continuous contact states.
This allows the centroidal controller to generate feedforward torques that effectively counteract ground reaction wrenches.
Experimental results on the Booster T1 humanoid demonstrated that HybridMimic achieved a 13\% reduction in base position tracking error compared to a state-of-the-art RL baseline during sim-to-real transfer.
The synergy between centroidal control and RL facilitates deterministic parameter tuning, such as adjusting velocity tracking gains, which streamlines the deployment process.
Finally, HybridMimic permits reduced joint stiffness, yielding compliant behavior for safer hardware operation.

% Despite its performance, HybridMimic is not without limitations. 
% Currently, the centroidal controller primarily generates feedforward torques to counteract the ground reaction wrench, focusing on the centroidal tracking accuracy.
% Consequently, the framework lacks an explicit physics-based mechanism for swing-leg dynamics, which is critical for precise foot placement and disturbance rejection during the swing phase. 
% Future research will explore the integration of limb-aware structures, such as task-space control, directly into the hybrid framework. 
% This evolution aims to unify centroidal dynamics with precise limb coordination, further reducing the tracking error of swing-phase trajectories and improving the robustness to external disturbances.

Although HybridMimic effectively regulates ground reaction forces for stability and performance, the current feedforward formulation does not explicitly optimize swing-leg trajectories, which are critical for precise foot placement. Future work will explore integrating model-based swing-leg control, such as task-space control, into the existing architecture to enhance the tracking accuracy during leg-swing phases of highly dynamic maneuvers.

% While we focus on motion mimicking tasks in this work, this framework could be easily extended to other RL pipelines. Additionally, the end-effector formulation allows this controller to be extended to robots with any number of limbs such as quadrupeds.

% However, our work on HybridMimic is not without limitations. 
% The centroidal controller mostly generate feedforward torque to counteract the ground reaction wrench, which does not explicitly benefit accurate swing leg tracking that is essential for foot placement.
% Future works would like to integrate a model-based structure that explicitly considers the swing foot position control, e.g., inverse kinematics control, in to current framework to further improve the control accuracy for swing limbs.
% In addition, the proper recipe of domain randomization for the centroidal-model-based controller requires more studies to improve robustness and to balance the effect from centroidal controller and PD controller.
% Future works include an empirical study of using HybridMimic to enable complex motion with multiple contact points not limited to legs and the study on the proper recipe of domain randomization for the centoridal-model-based controller.
% In the future, an empirical study using the HybridMimic framework to mimic motions with contact involving hands and other contact surfaces will be conducted to further demonstrate the potential of the proposed method, enabling complex motions with multiple contacts. 

\section*{Acknowledgment}
The authors would like to thank Muqun Hu and Zijian He for insightful discussion and thank Zenan Zhu for experimental support. 
Hardware support from Booster Robotics is gratefully acknowledged.
This work used Jetstream2 GPU at Indiana University through allocation MCH250079 from the Advanced Cyberinfrastructure Coordination Ecosystem: Services \& Support (ACCESS) program, which is supported by U.S. National Science Foundation grants \#2138259, \#2138286, \#2138307, \#2137603, and \#2138296. 
The data used in this project was obtained from mocap.cs.cmu.edu.
The database was created with funding from NSF EIA-0196217.

\bibliographystyle{IEEEtran}
\bibliography{Reference}
\end{document}